\documentclass[a4paper, 10pt, conference]{ieeeconf}      

\IEEEoverridecommandlockouts                              

\usepackage{cite}
\usepackage{amsmath,amssymb,amsfonts}
\usepackage[linesnumbered,lined,ruled,commentsnumbered]{algorithm2e}
\usepackage{graphicx}
\usepackage{textcomp}
\usepackage{subfig}
\usepackage{hyperref}
\usepackage{array}

\usepackage{gensymb} 

\title{\LARGE \bf
The Autonomous Siemens Tram
}

\author{Andrew W. Palmer, Albi Sema, Wolfram Martens, Peter Rudolph, and Wolfgang Waizenegger
\thanks{The authors are with Siemens Mobility GmbH, Berlin, Germany}%
\thanks{Corresponding author: {\tt\small andrew.palmer@siemens.com}}%
}

\begin{document}

\begin{table*}
\copyright 2020 IEEE. Personal use of this material is permitted. Permission from IEEE must be obtained for all other uses, in any current or future media, including reprinting/republishing this material for advertising or promotional purposes, creating new collective works, for resale or redistribution to servers or lists, or reuse of any copyrighted component of this work in other works.

The published version of this article can be found at https://doi.org/10.1109/ITSC45102.2020.9294699
\end{table*}

\pagebreak

\maketitle
\thispagestyle{empty}
\pagestyle{empty}

\begin{abstract}

This paper presents the Autonomous Siemens Tram that was publicly demonstrated in Potsdam, Germany during the InnoTrans 2018 exhibition. The system was built on a Siemens Combino tram and used a multi-modal sensor suite to localize the vehicle, and to detect and respond to traffic signals and obstacles. An overview of the hardware and the developed localization, signal handling, and obstacle handling components is presented, along with a summary of their performance. 


\end{abstract}

\section{Introduction}

Autonomous driving in the rail industry is in its infancy. The Autonomous Siemens Tram (AST) project was an opportunity to investigate the applicability of autonomous driving technologies developed by the automotive industry to the rail domain, and demonstrate what capabilities a future autonomous tram may offer. In many respects, autonomous driving for trams is very similar to autonomous driving for cars---they operate in similar environments where they interact with other road users such as cars, pedestrians, and cyclists, and they must obey similar traffic rules and signals. Some aspects of the problem are simplified by the rail-bound nature of the vehicles---there are limited areas that need to be mapped, path planning is not required, and the possible locations of the vehicle are heavily restricted. However, the rail-bound nature also makes the problem of avoiding obstacles considerably more challenging. Not only are trams unable to steer in order to avoid potential collisions, they also cannot decelerate as fast as cars, both due to physical limitations and the risk of injuring unsecured passengers. 

\begin{figure}[]
    \centering
    \includegraphics[width=\linewidth]{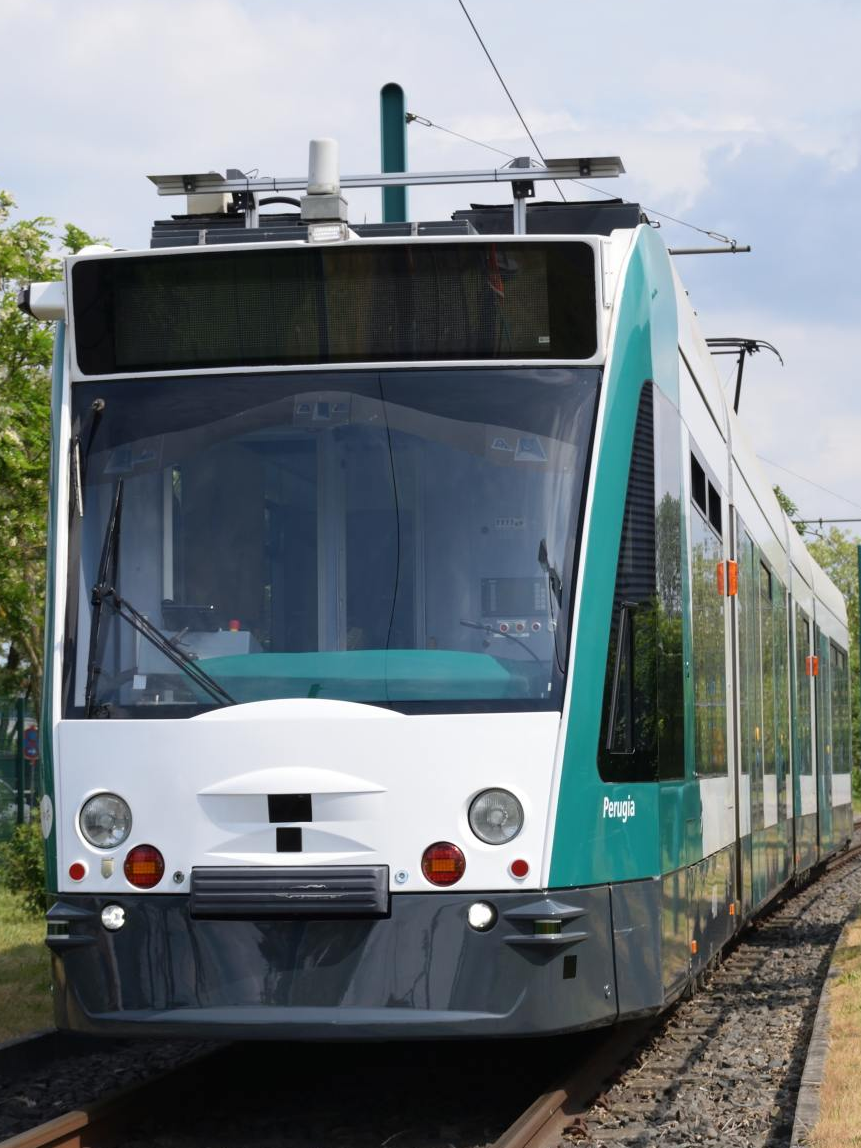}
    \caption{The Autonomous Siemens Tram}
    \label{f:tram}
\end{figure}

\begin{figure*}[]
	\centering
	\includegraphics[width=\linewidth]{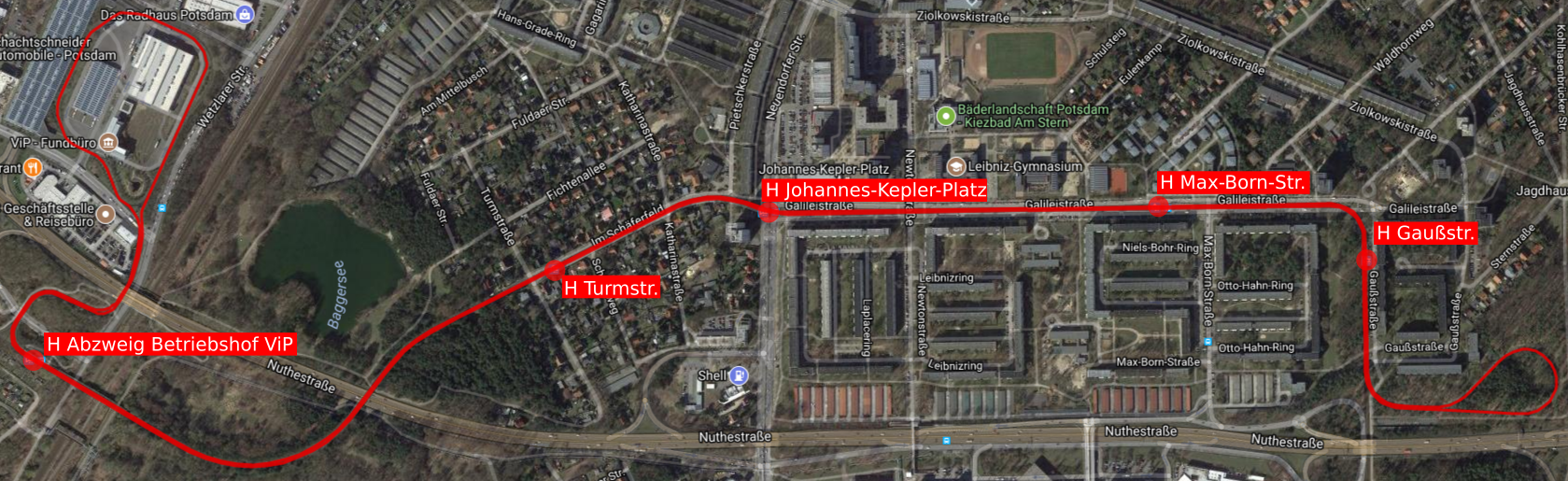}
	\caption{Map of the route with station locations marked}
	\label{f:map}
\end{figure*}

The AST, shown in Figure~\ref{f:tram}, was developed during 2018 and publicly demonstrated in Potsdam, Germany during the InnoTrans 2018 exhibition, where it performed successful demonstration drives for hundreds of passengers. Note that a safety driver was always present during autonomous operation. It was demonstrated on a 6km long section of the Potsdam tram network shown in Figure~\ref{f:map}. This track was comprised of a number of different environments that presented many of the scenarios that trams generally encounter. Between the stations `H Abzweig Betriebshof ViP' and `H Turmstr.', the track is surrounded by heavily wooded areas where it was not uncommon to see an animal crossing the tracks. Around `H Turmstr.' is a suburban area with several unsignalised road crossings and low fences surrounding the track in between the crossings. The section of track around the stations `H Johannes-Kepler-Platz' and `H Max-Born-Str.' has higher density buildings, with businesses, schools, and apartment buildings on both sides of the track. There are a number of signalised and unsignalised crossings in this area, and, between the crossings, there is a fence in between the tracks to encourage pedestrians to cross at the designated crossings. After the station `H. Gau{\ss}str.' there is a return loop through another wooded area where the tram began the return journey. Electrical power was supplied to the vehicle via overhead catenary lines, with the poles positioned in between the two parallel tracks. 

This paper presents an overview of the hardware and software comprising the AST and a summary of their performance. The hardware is first introduced in Section~\ref{s:hardware}, with the software architecture, algorithms, and performance following in Section~\ref{s:software}. Section~\ref{s:conclusion} concludes and presents possible avenues of improvement. 

\section{Hardware Overview}\label{s:hardware}

This section presents an overview of the vehicle, computing hardware, and sensors used in the project. 

\subsection{Vehicle}

The vehicle used, shown in Figure~\ref{f:tram}, was the prototype Siemens Combino NF100 low floor tram built in 1996. It is approximately 26m long, 3.5m high, and 2.3m wide, and runs on standard gauge track (1.435m gauge). It has an unladen weight of approximately 28 tonnes, and is rated for 150 passengers. This corresponds to an additional load of approximately 12 tonnes, or 40\% of the unladen weight. The vehicle has a theoretical maximum speed of 70km/h, but on the Potsdam track network it is limited to 50km/h. Compared to regular road vehicles, its performance is very limited---it has a maximum acceleration of 1.3m/s$^2$, an average braking deceleration when using the service brake of approximately 1.2m/s$^2$, and with the emergency brake of up to 3m/s$^2$. The emergency brake, however, is rarely used, as unsecured passengers are likely to be injured during its operation. Consequently, the emergency brake was not used by the AST and its operations was the responsibility of the safety driver. 

\subsection{Computers} \label{s:computers}

A number of computers were used to perform the various parts of the autonomous operations of the vehicle. Several computers equipped with graphics cards were used to perform camera-based signal recognition and object detection tasks, and a number of railway-certified computers were used for localization, lidar- and radar-based object detection, and object fusion. All computers were interconnected via Ethernet. 
In addition to these computers, a tablet was used to display the current state of the system to the safety driver.


\subsection{Sensors}

The following subsections introduce the multi-modal suite of sensors that the vehicle was equipped with to perform the localization, signal handling, and obstacle handling tasks. 

\begin{figure}[]
	\centering
	\subfloat[Lidars]{
		\includegraphics[width=0.47\linewidth]{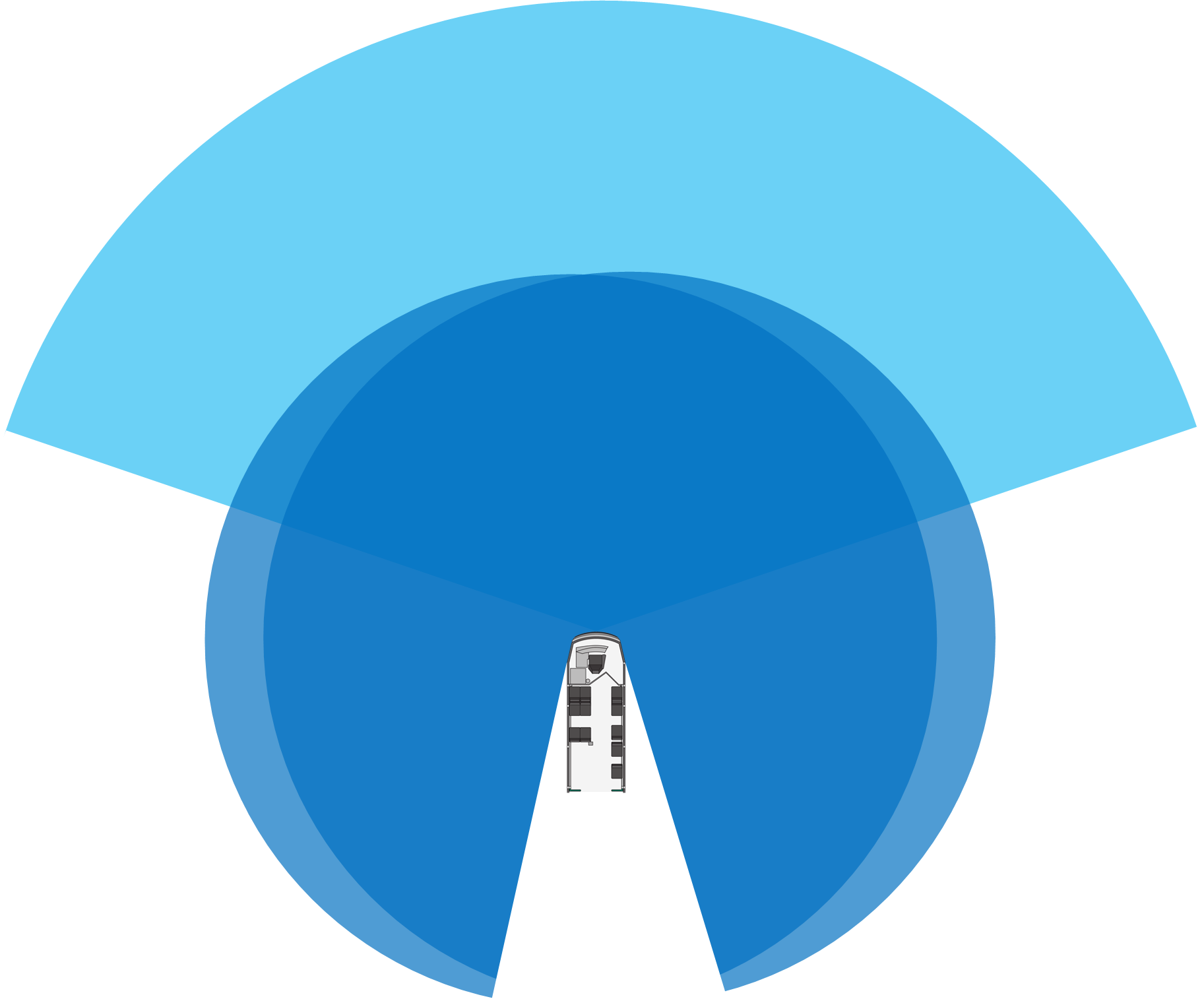}\label{f:sensors_lidar}
	}
	\subfloat[Radars]{
		\includegraphics[width=0.47\linewidth]{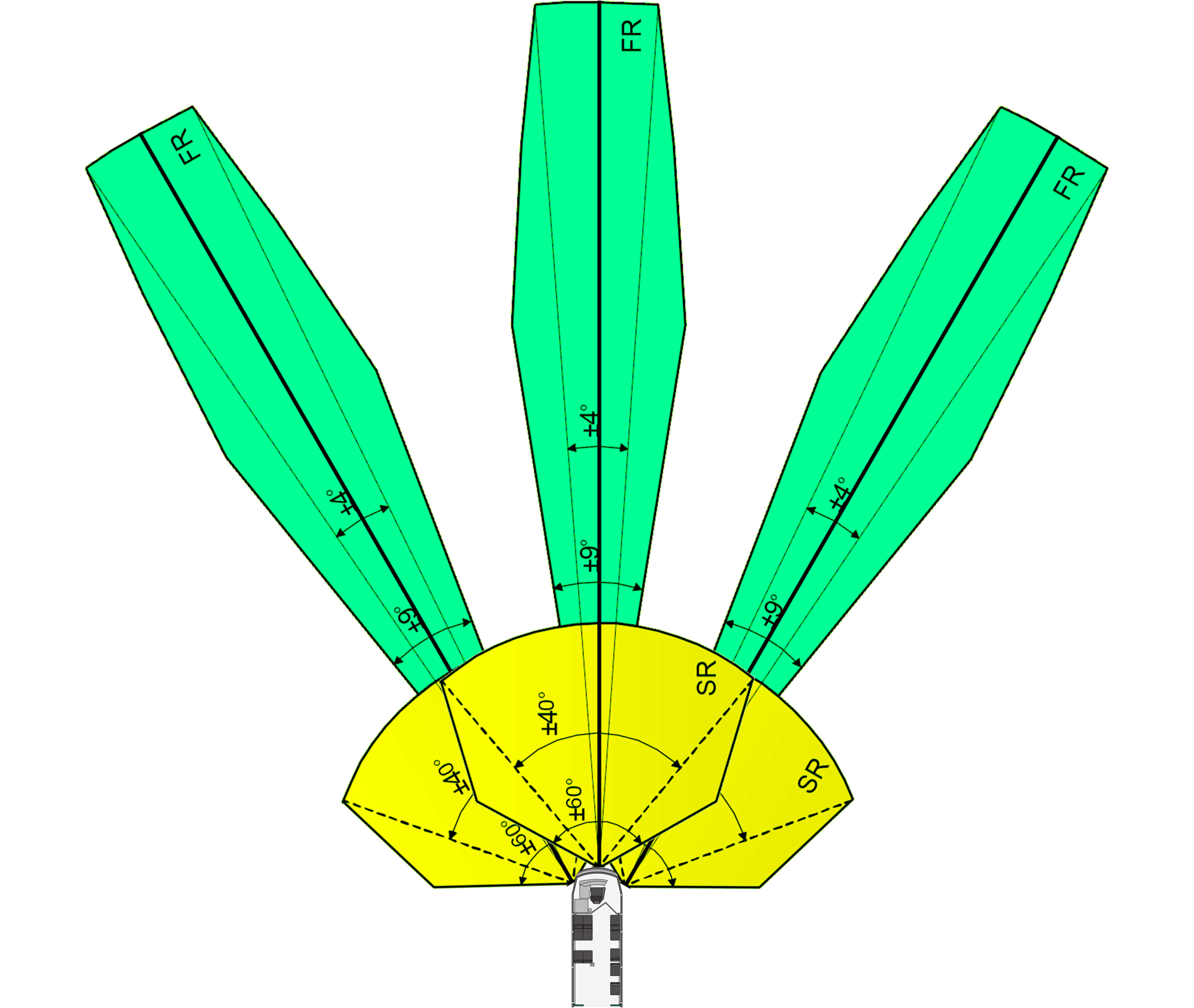}\label{f:sensors_radar}
	}
	
	\subfloat[Object detection cameras]{
		\includegraphics[width=0.47\linewidth]{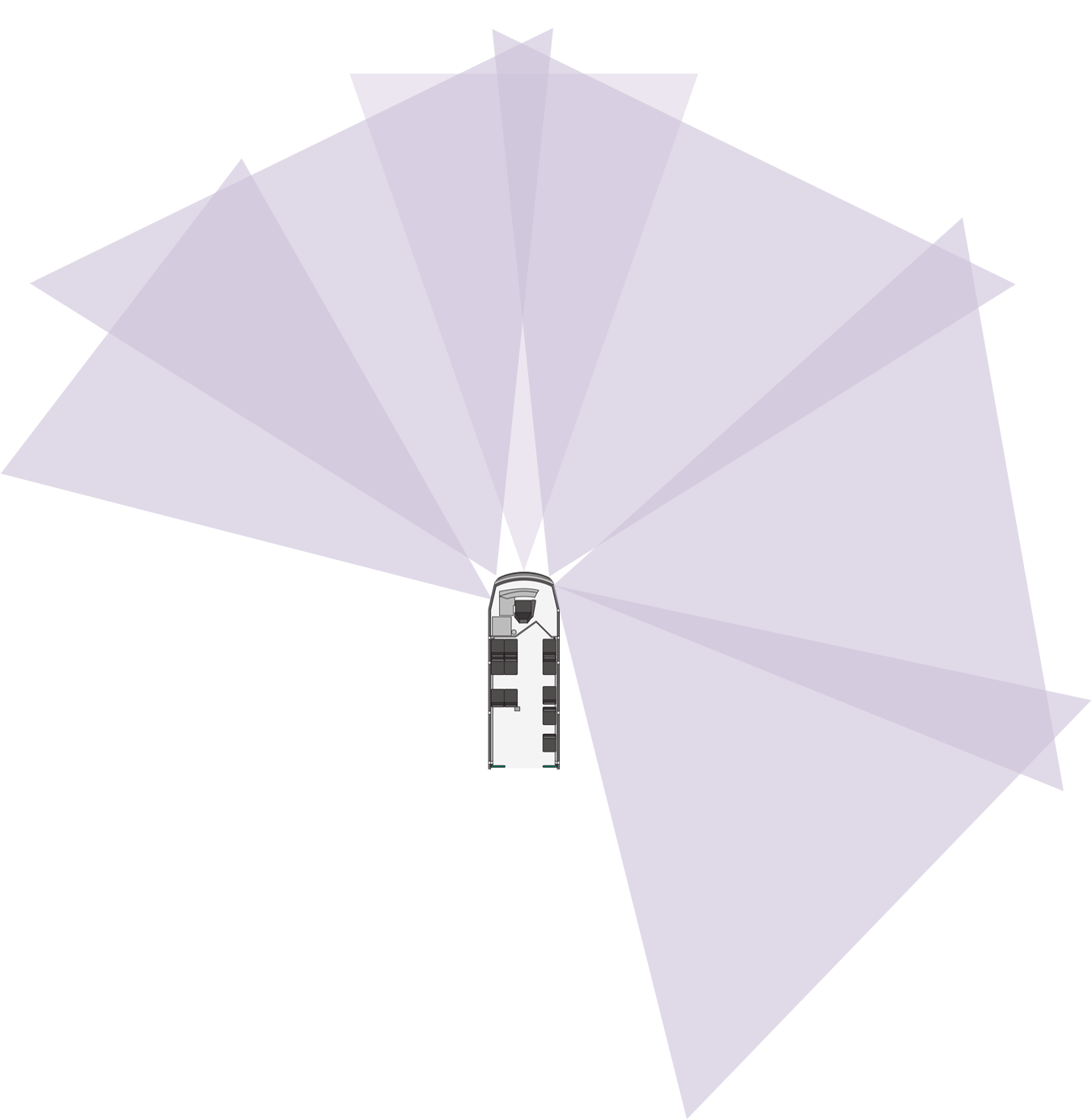}\label{f:sensors_sekonix}
	}
	\subfloat[Signal recognition cameras]{
		\includegraphics[width=0.47\linewidth]{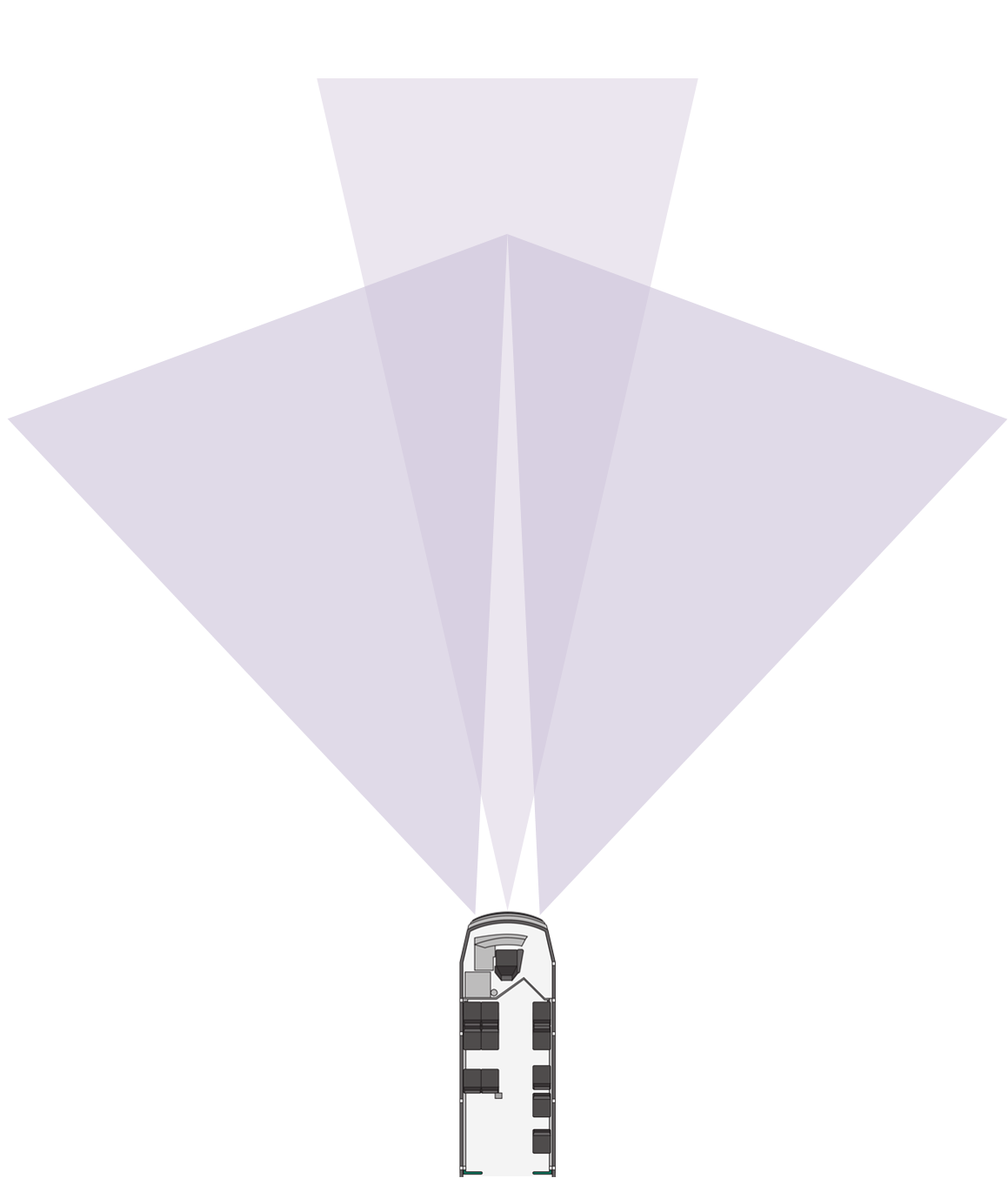}\label{f:sensors_basler}
	}
		
	\caption{Sensor field of view}
	\label{f:sensors}
\end{figure}

\subsubsection{Localization}

A dual antenna GNSS-aided Inertial Navigation System (INS) was used to provide highly accurate pose and velocity information. Combined with a Real-Time Kinematic (RTK) correction data service, the INS was capable of providing measurements at 100Hz with up to 0.01m position accuracy, 0.05km/h velocity accuracy, 0.03\degree~roll and pitch accuracy, and 0.1\degree~heading accuracy. The 2 antennas were mounted above the roof of the vehicle with a separation of 1.6m. 

\subsubsection{Lidar}

Two separate types of lidars were used on the vehicle---a forward-facing 4-layer automotive laser scanner with integrated object tracking was used for long-range object perception, while two 16-layer lidars were mounted on the front corners of the vehicle to provide a wide field of view for near-range object perception.  The field of view of the lidars is shown in Figure~\ref{f:sensors_lidar}. 

\subsubsection{Radar}

A set of 3 automotive radars were mounted at the front of the vehicle at various angles. The field of view of the radars is shown in Figure~\ref{f:sensors_radar}. The radars have two separate fields of view shown in yellow (up to 70m) and green (up to 250m) respectively. Similar to the forward-facing lidar, the radar has integrated object tracking and is capable of tracking up to 100 objects at a time. 


\subsubsection{Camera}

The vehicle was equipped with a total of 9 cameras. A set of 6 cameras were positioned behind the windshield and side windows in the front cab of the vehicle to provide a wide field of perception for the purpose of object detection, as shown in Figure~\ref{f:sensors_sekonix}. The lens on the camera facing directly forwards had a 60\degree~Field of View (FoV), and the other 5 cameras each had a 120\degree~FoV. The placement of the cameras on the right side of the vehicle was chosen in order to provide perception of pedestrians who may walk in front of the tram from the platform while the tram is stationary, and of vehicles travelling parallel on the road to the right of the track who present a particularly high risk of collisions at road crossings when they turn left across the track. 
A further set of 3 cameras were used to provide forward facing perception for signal recognition. They were positioned behind the windshield of the vehicle as shown in Figure~\ref{f:sensors_basler}. The left and right cameras cameras had a 60\degree~FoV, while the centre camera had a 35\degree~FoV. 



\section{Software Overview}\label{s:software}

\begin{figure}[]
	\centering
	\includegraphics[width=\linewidth]{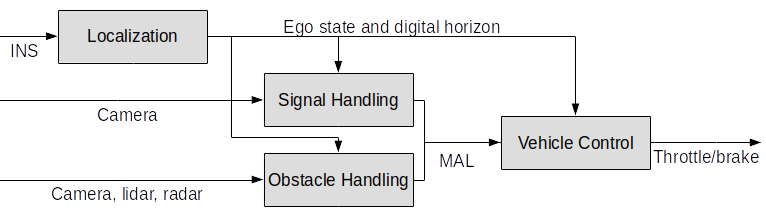}
	\caption{Software architecture}
	\label{f:architecture}
\end{figure}

A high-level overview of the software architecture used in the AST is shown in Figure~\ref{f:architecture}. It consisted of four main subsystems---vehicle control, localization, signal handling, and obstacle handling---each of which are summarised in the following sections. 

\subsection{Vehicle Control}

Vehicle control was provided by a proprietary system. Using the position and speed of the vehicle on the map, it determines the throttle and brake commands necessary to drive from platform to platform while obeying speed limits. A Movement Authority Limit (MAL) input is additionally used to specify the distance that the vehicle is allowed to move. The MAL input is used to respond to the dynamic environment, and is calculated by taking the minimum of the MAL outputs of the signal handling and obstacle handling components. 


\subsection{Localization Subsystem}

The localization subsystem used the INS in combination with a map of the track and associated infrastructure to provide state information to the vehicle control, as well as a digital horizon to the signal and obstacle handling subsystems. The digital horizon provided information about the upcoming track including the track geometry, signals, platforms, and crossings. The following subsections present details on how the map was created, and how the localization was performed. 

\subsubsection{Maps}


The first step in creating the map was to convert position data collected using the INS into the track network. The raw INS data was post-processed to minimise any effects from poor GNSS signal. The number of points in the resultant trajectory were reduced primarily by automatically identifying straight sections of track and removing unnecessary points on these sections. Infrastructure was then mapped by hand, including the start and end of each platform, the stopping point of the vehicle at each platform, the signals (including their height and type), the start and end of each road and pedestrian crossing, and the electrical grid separators (which are locations where the vehicle is not allowed to stop as it is where there is a physical gap in the overhead power supply). Finally, any other additional information, such as speed limits, were added manually. 

\subsubsection{Localization}

In many railway applications, localization can be very challenging \cite{Lauer2015,Palmer2018}, and difficulties in particular arise from ambiguity over which track the vehicle is on. This can be reduced to a 1-dimensional problem using information from the operation centre about which track is allocated for the current trip. This, combined with the INS, was sufficient for providing the location of the vehicle on the track. 



\subsubsection{Performance}

The performance of the INS was excellent in open sky conditions, providing a reported position accuracy of a few centimetres. However, in certain sections of the track the view of the sky was partially obscured by overhanging vegetation and tall buildings. In these cases, any degradation in GNSS signal was overcome using dead reckoning based on the IMU and motion model. 
It was observed that, during longer periods without GNSS (for example, inside depot buildings), the drift of the INS was higher than would normally be expected. It is hypothesised that this was due to the INS using a sophisticated motion model based on car motion which does not accurately reflect the motion of a rail-bound vehicle. 



\subsection{Signal Handling Subsystem}

The signal handling subsystem consisted of three components---signal state detection, signal state filtering, and a signal planner. The signal state detection component identified the state of a signal in a single camera image. As the aim of this project was to use only on-board technologies without modifying existing infrastructure, technologies such as V2X were not adopted for signal state identification. Detections from multiple cameras and frames were then combined and filtered in the signal state filtering component, with the resultant filtered signal state being used by the signal planner to determine the MAL necessary to obey the signal. The following subsections present an overview of each component, and a summary of the performance of the subsystem. 

\subsubsection{Signal State Detection}

\begin{figure*}[]
	\centering
	\subfloat[A]{
		\includegraphics[width=0.095\linewidth]{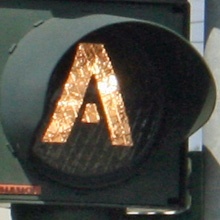}\label{sf:A}
	}
	\subfloat[F0]{
		\includegraphics[width=0.095\linewidth]{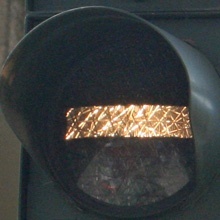}\label{sf:F0}
	}
	\subfloat[F1]{
		\includegraphics[width=0.095\linewidth]{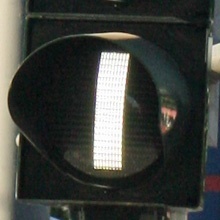}\label{sf:F1}
	}
	\subfloat[F2]{
		\includegraphics[width=0.095\linewidth]{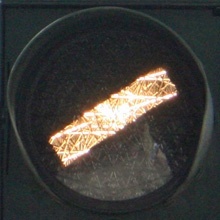}\label{sf:F2}
	}
	\subfloat[F3]{
		\includegraphics[width=0.095\linewidth]{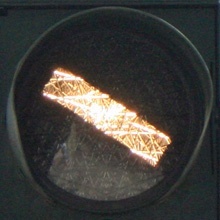}\label{sf:F3}
	}
	\subfloat[F4]{
		\includegraphics[width=0.095\linewidth]{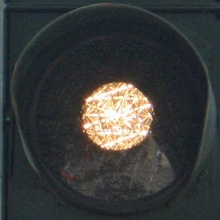}\label{sf:F4}
	}
	\subfloat[W0]{
		\includegraphics[width=0.095\linewidth]{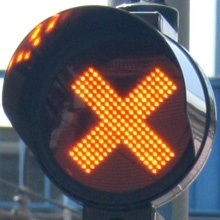}\label{sf:W0}
	}
	\subfloat[W12]{
		\includegraphics[width=0.095\linewidth]{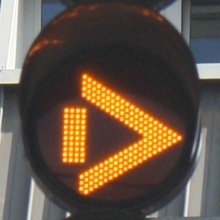}\label{sf:W12}
	}
	\subfloat[W13]{
		\includegraphics[width=0.095\linewidth]{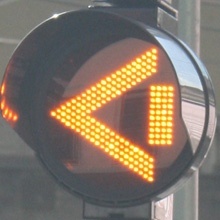}\label{sf:W13}
	}
	
	\caption{The possible signal states along the test track: (a)~Approaching tram has been registered by the signal (b)~Stop (c)~Go straight (d)~Go right (e)~Go left (f)~Get ready (g)~Switch state locked (h)~Right branch selected (i)~Left branch selected}
	\label{f:signals}
\end{figure*}

German tram signals are quite different to regular traffic signals. There are two categories of signals---stop-go signals (which correspond in functionality to regular red-yellow-green traffic signals), and switch state signals (which indicate which direction a railway switch is set to). Stop-go signals, shown in Figures~\ref{sf:A}-\ref{sf:F4}, will always contain an F0 (stop) and an F4 (get ready) signal, and an F1, F2, or F3 go signal which also indicates the track direction that it applies to (in the case of a branching track). 
In many cases, they will also include an A signal (which stands for \textit{Anforderung} in German) which, when lit, means that the approaching tram has been registered by the signal and the upcoming light sequence will include a go signal for the tram. 
Switch signals, shown in Figures~\ref{sf:W0}-\ref{sf:W13}, contain a W0 signal, which indicates that the switch state is locked for an approaching vehicle, and W12 and W13 signals specifying the turning direction. 

As there is no flow of information from the signal to the tram about its state, the signal state is detected using camera images. 
Multiple approaches in the automotive industry determine the state of a signal by detecting and classifying the whole signal housing \cite{jensen2016vision}. In comparison to traffic signals, tram signals can have varying numbers of chambers, each indicating one of the states mentioned above, giving many possible combinations that need to be detected. As a result, methods that detect signals as a whole were not pursued. The approach that was developed detects the state of each chamber separately, and the combination of the detected chambers along with prior knowledge from the digital horizon enables the overall signal state to be resolved. 

\begin{figure}[]
    \includegraphics[width=\linewidth]{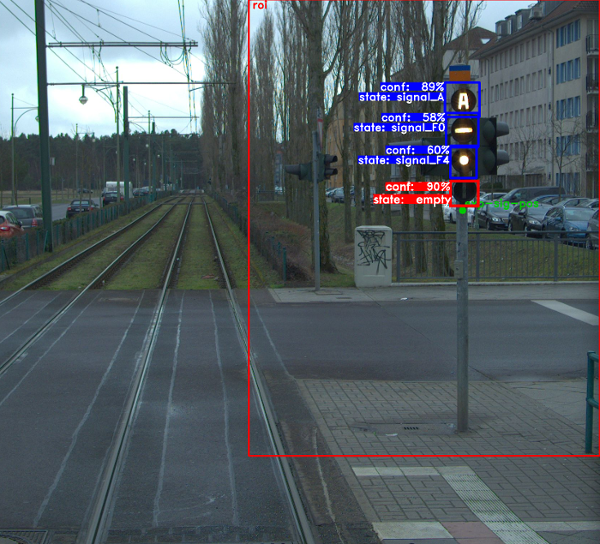} 
    \caption{Example detection of signal chambers. The expected signal location from the digital horizon is shown in green, the large red box is the ROI generated from the expected location, and the small boxes show the individual detections of lit chambers (blue) and empty chambers (red).  }
    \label{f:sgn_detection}
\end{figure}

The family of Single Shot Detectors (SSD) provide a good trade-off between accuracy and speed, making them suitable for real-time systems. Through multiple experiments, it was determined that the best combination of good real-time performance and accurate detections of small signals at far distances could be achieved using a modified version of the original SSD architecture \cite{DBLP:journals/corr/LiuAESR15}. 
This modified network used a MobileNet v2 feature extractor \cite{DBLP:journals/corr/abs-1801-04381} with an input size of 512x512. The default size of the anchors was set to a square ratio of 1:1 with smaller default dimensions than the original network, fitting to the scope of the problem. 
The location of the signal in the digital horizon was used to identify a Region of Interest (ROI) in the image where the signal was expected to be. 
Using the ROI as the input to the network made the signal chambers a significant enough size to be accurately recognized by a fast SSD algorithm. An example result of the algorithm is shown in Figure~\ref{f:sgn_detection}.

The network was trained on the labels shown in Figure~\ref{f:signals}. One of the challenges that was encountered was that the shape of the signal in each chamber was partially visible, even if the chamber itself was not lit---this can be observed in the bottom chamber in Figure~\ref{f:sgn_detection}. To prevent these from being classified as a lit signal chamber, an extra ``empty" label was added to the training set to represent the chambers that are not lit. 
Training of the detection network was done on a manually labelled dataset of 10,000 images, of which 1,000 random images were chosen for validation during training. Image augmentation was also applied during training, excluding rotation and mirroring. A further 700 hand-picked images from multiple challenging scenarios and weather situations were used as the test set.

\subsubsection{Signal State Filtering}

A Bayes filter was used to track the state of a signal. As each chamber of a signal was detected separately, a Hidden Markov Model (HMM) was used to describe which combinations of chambers could be active at any one time, and the possible transitions between states, in order to reduce misclassifications. 
Before integrating the detections into the filter, implausible detections were identified and discarded using the expected size and possible signal states as indicated in the digital horizon. 


\subsubsection{Signal Planning}

A stopping location was determined for each signal that ensured that the signal would always be in the field of view of at least one of the cameras when the vehicle came to a stop. 
When an upcoming signal was indicated in the digital horizon, the MAL was set to the stopping point of the signal until the necessary combination of activated chambers was identified in the filtered signal state to allow the vehicle to proceed past the signal. This meant that it would always stop in front of the signal if the signal detection or filtering failed to detect the signal state. 


A second location was also specified for each signal which was the position at which the vehicle, if it could not stop by that point, would continue to drive past the signal. This was necessary to prevent the vehicle from braking in situations where the signal changed to a prepare to stop state (F4), but the braking distance of the vehicle was greater than the distance to the point where none of the cameras would be able to detect the signal state. 


\subsubsection{Performance}

The performance of the signal detection on the camera images was measured by using object detections metrics such as mean Average Precision (mAP), mean Average Recall (mAR), and custom metrics such as the maximum distance at which the state could be accurately detected. Table~\ref{t:signal_results} shows the performance metrics for the developed network on three of the most challenging weather categories in the test set---cloudy, raining, and sunny with reflections.

\begin{table}[]
\centering
\caption{Signal detection results}
\begin{tabular}{ | c || c | c | c | } 
\hline
\textbf{Dataset} & \textbf{mAP} & \textbf{mAR} & \textbf{Max. detection distance} \\ 
\hline
Cloudy & 0.65 & 0.75 & 80-100m \\ 
\hline
Raining & 0.6 & 0.73 & 80m \\ 
\hline
Sunny w/ reflections & 0.49 & 0.48 & 70-80m \\ 
\hline
\end{tabular}
\label{t:signal_results}
\end{table}

During operation, the passive behaviour of always planning on stopping at a signal unless it was positively identified ensured that the vehicle never drove past a signal where it should stop. Missed and incorrect detections were filtered out, and the end behaviour of the system was robust, even in poor weather conditions. 

\subsection{Obstacle Handling Subsystem}

The goal of the obstacle handling subsystem was to detect and respond to dynamic objects in the environment. Two different approaches were implemented for detecting possible obstacles---an object-based approach, in which objects were detected in multiple sensor modalities and combined using a late fusion approach, and a free space detection approach which worked purely with lidar sensors. 
An obstacle planner used the detected objects and free space, along with the digital horizon, to determine the MAL necessary to avoid obstacles on the track and, additionally, whether the warning bell should be activated. The following subsections detail the two obstacle detection approaches, along with the obstacle planner, and the overall performance of the obstacle handling subsystem.  

\addtolength{\textheight}{-1.2cm}   

\subsubsection{Object Detection and Filtering} \label{s:object}

The object detection and filtering approach utilised the full multi-modal sensor suite consisting of cameras, lidars, and radars. A commercially available deep learning-based object detector developed for automotive applications was used for detecting and classifying objects from the camera data, while the automotive lidar and radars provided classified object outputs out of the box. These detections were combined using a Kalman filter, where measurements were associated to objects using Mahalanobis distance and nearest-neighbour assignment. The success of such an approach is heavily dependent on the quality of the object detections. In this project, sensors and object detection algorithms designed for the automotive industry were used with the hypothesis that, given trams operate in a similar environment to road vehicles, the performance should be similar. However, the tram environment introduced a number of challenges when using automotive sensors. 

The largest challenge for using the object detections from the radars was that the strongest returns typically come from objects with lots of metal. As most of the infrastructure on and around the track has a very high metal content, the most commonly detected objects by the radar were the power poles and the railway sleepers. Returns from pedestrians were indistinguishable amongst the infrastructure detections. Another problem encountered was that, unlike for other sensor modalities, large objects were typically detected as multiple smaller objects---this was especially the case for fences and other trams. 

Similar to the radar, the majority of detections from the lidar were also of the infrastructure. The high density of infrastructure objects (in particular, fences and power poles) again posed a significant challenge for the object fusion, as object detections that were close together would be associated with one another, leading to non-zero velocities. As the infrastructure is typically very close to the track, any imprecision in the position of an object generated from a piece of infrastructure can lead to a collision being predicted. 

The object detections from the cameras were, in general, quite good, but were sensitive to the position of the camera---when using cameras that were positioned high on the windscreen, objects that were close to the tram were sometimes not detected. Despite this, cars and pedestrians were almost always detected. However, the false positive rate was also high. 
Integrating the object detections into the object fusion was particularly challenging. As the detections were made in image space, the projection of the object detection to 3D relied on very accurate calibration of the sensors and accurate bounding boxes---estimating the position of partially occluded objects was especially difficult. 


Predicting the future trajectory of an object is a challenging topic, and reliable results are achievable for a few seconds \cite{Hoermann2018}. In this case, the large braking distances of trams limit the usefulness of prediction, as the large braking distances can necessitate a similarly long prediction horizon.  

\addtolength{\textheight}{-0.5cm}

\subsubsection{Free Space Detection}

A free space-based approach was also implemented to provide redundancy in the obstacle handling. It combined the raw point clouds of the lidars in order to determine which areas of the environment were free or occupied space using the following process. The latest point clouds from each sensor were first spatially and temporally aligned. 
Then, irrelevant points were filtered out considering the clearance gauge of the vehicle. 
Following this, the remaining points were clustered, projected to the ground plane, and converted to polygons---these polygons represented the occupied space in the environment. 

A more conservative approach was also implemented, in which areas that were not visible due to occlusions were explicitly modelled as occupied space. While such an approach is safer than using the convex hull of each cluster, in practice it led to undesirable obstacle detections when approaching curves, where pieces of infrastructure would partially occlude the track. 

A challenging problem that was encountered was vegetation growing on the track. Unlike in the automotive domain where the road surface is essentially flat, the ground surface between and around the rails was sometimes loose gravel and could have plants growing in it.  
Choosing the correct clearance gauge was a trade-off between being able to detect objects close to the ground and not detecting tall vegetation. 

\subsubsection{Obstacle Planner}

The obstacle planner calculated whether collision and warning detection zones defined for each part of the track intersected with the objects and occupied space polygons in order to determine the actions to be taken. Default collision and warning detection zones were used for the majority of the track, with special zones defined for high-risk areas such as platforms and crossings. For platforms in particular, pedestrians tend to stand very close to the track and should be warned when the tram is approaching. The basic response strategy was to set the MAL to a specified offset in front of a detected collision. If the distance to the obstacle was less than the braking distance (i.e., the tram was predicted to collide with the obstacle) or the obstacle was within a predefined distance, then the warning bell would additionally be used. Activating the warning bell was the only response for obstacles detected in a warning zone. 

\subsubsection{Performance}

To evaluate the performance of the obstacle handling, a pram was used as a test obstacle---this was able to be reliably detected over 80m away. As the braking distance of the vehicle at the maximum velocity of 50km/hr was approximately 80m, the velocity of the vehicle when approaching pedestrian crossings was limited during operation to 40km/hr to provide a safety buffer. 
As this approach projects the detected objects and free space onto the map, it is very sensitive to errors in sensor-to-vehicle calibration, and errors in the estimated vehicle position and orientation. With poles and fences positioned very close to the track, even small angular errors can lead to false detections. Sensor-to-sensor calibration was performed using a set of calibration targets placed around the vehicle, with the sensor-to-vehicle manually estimated using a straight section of track as the reference. In combination with the high accuracy of the localization, the alignment of the sensor data with the track map was sufficient and false collision detections occurred very rarely. 

\section{Conclusion}\label{s:conclusion}

This paper presented a summary of the Autonomous Siemens Tram, with a focus on the software developed and the challenges that were encountered. As was shown, the application of automotive sensors and technologies to the railway domain is not straight forward and requires significant adaptation for the unique situations that these vehicles encounter. Despite successful demonstrations of the system, there are a number of possible avenues for improving the robustness of the system. It is already clear that GNSS-based solutions for localization will not be sufficient for reliable operation in all environments. Research is underway on using visual odometry in the rail domain \cite{Tschopp2019}, and perception-based localization approaches are likely to be necessary to fully cover the situations where GNSS-based solutions fail. Object-based obstacle handling could also be improved by using a map of the infrastructure to determine whether a measurement is likely to be of a piece of infrastructure and can be discarded. However, the challenge with such an approach is ensuring that dynamic objects positioned next to infrastructure are not erroneously discarded.









\section*{Acknowledgment}

Our thanks go to the large team of Siemens Mobility employees who contributed to the success of this project. We would also like to acknowledge the support of the Verkehrsbetrieb Potsdam GmbH who provided the vehicle, safety drivers, and access to the track.


\bibliographystyle{IEEEtran}
\bibliography{IEEEabrv,ast_bib}




\end{document}